\documentclass{article}
\usepackage[toc,page]{appendix}

\usepackage{PRIMEarxiv}
\usepackage{todonotes}
\usepackage[utf8]{inputenc} % allow utf-8 input
\usepackage[T1]{fontenc}    % use 8-bit T1 fonts
\usepackage{hyperref}       % hyperlinks
\usepackage{url}            % simple URL typesetting
\usepackage{booktabs}       % professional-quality tables
\usepackage{amsfonts}       % blackboard math symbols
\usepackage{nicefrac}       % compact symbols for 1/2, etc.
\usepackage{pdflscape}
\usepackage{microtype}      % microtypography
\usepackage{lipsum}
\usepackage{fancyhdr}       % header
\usepackage{float}
\usepackage{graphicx}       % graphics
\usepackage{diagbox}
\graphicspath{{media/}}     % organize your images and other figures under media/ folder
\usepackage{multirow}
\usepackage{amsmath}
\usepackage{tikz}
\usetikzlibrary{shapes.geometric, arrows}

\tikzstyle{sq} = [rectangle, rounded corners, minimum width=2cm, minimum height=1cm,text centered, draw=black, fill=red!20,text width=2cm,align=center]
\tikzstyle{process} = [rectangle, rounded corners, minimum width=2cm, minimum height=1cm,text centered, draw=black, fill=blue!20,text width=3cm,align=center]
\tikzstyle{process2} = [rectangle, rounded corners, minimum width=2cm, minimum height=1cm,text centered, draw=black, fill=blue!20,text width=2cm,align=center]
\tikzstyle{decision} = [diamond, rounded corners, minimum width=1cm, minimum height=1cm,text centered, draw=black, fill=green!20,text width=2cm,align=center]
\tikzstyle{arrow} = [thick,->,>=stealth]

\title{Automated Multi-Label Annotation for Mental Health Illnesses Using Large Language Models}

\author{ Abdelrahaman A. Hassan$^\dagger$,  Radwa J. Hanafy$^{\dagger,\ddagger}$ and  Mohammed E. Fouda$^\dagger$ \\
$^\dagger$Compumacy for Artificial Intelligence Solutions, Cairo, Egypt.\\
$^\ddagger$ Department of Behavioural Health- Saint Elizabeths Hospital, Washington DC, 20032.\\
  \texttt{fouda@compumacy.com} 
}

\begin{document}
\maketitle

\begin{abstract}

The growing prevalence and complexity of mental health disorders present significant challenges for accurate diagnosis and treatment, particularly in understanding the interplay between co-occurring conditions. Mental health disorders, such as depression and Anxiety, often co-occur, yet current datasets derived from social media posts typically focus on single-disorder labels, limiting their utility in comprehensive diagnostic analyses. This paper addresses this critical gap by proposing a novel methodology for cleaning, sampling, labeling, and combining data to create versatile multi-label datasets. Our approach introduces a synthetic labeling technique to transform single-label datasets into multi-label annotations, capturing the complexity of overlapping mental health conditions. To achieve this, two single-label datasets are first merged into a foundational multi-label dataset, enabling realistic analyses of co-occurring diagnoses. We then design and evaluate various prompting strategies for large language models (LLMs), ranging from single-label predictions to unrestricted prompts capable of detecting any present disorders. After rigorously assessing multiple LLMs and prompt configurations, the optimal combinations are identified and applied to label six additional single-disorder datasets from RMHD. The result is SPAADE-DR, a robust, multi-label dataset encompassing diverse mental health conditions. This research demonstrates the transformative potential of LLM-driven synthetic labeling in advancing mental health diagnostics from social media data, paving the way for more nuanced, data-driven insights into mental health care.
\end{abstract}

% \keywords{LLM \and Second keyword \and More}

\section{Introduction}

Mental illnesses are a major global health issue, affecting millions of people across all age groups and societies. Mental disorders not only impact the well-being and quality of life of individuals but also contribute to a significant burden on healthcare systems worldwide. Mental health conditions encompass a broad spectrum, including disorders like depression, anxiety, and PTSD, as well as more severe outcomes like suicide \cite{APA2013}.
For instance, depression affects approximately 280 million people globally, accounting for 3.8\% of the population. Among adults, 5\% are affected, with a higher prevalence in women (6\%) compared to men (4\%)\cite{who2023depression}. Anxiety disorders, the most common of all mental health conditions, impact around 301 million people globally, with 4\% of the world’s population experiencing these disorders\cite{who_anxiety}. Moreover, the severe consequences of untreated mental illness are starkly evident in the global suicide rate, where an estimated 726,000 people take their own lives each year, making it one of the leading causes of death, especially among young people aged 15 to 29\cite{who_mental_health_report}.These statistics illustrate the widespread nature and severity of mental illnesses, underscoring the urgent need for effective methods of early diagnosis and intervention.
In recent years, social media platforms have emerged as valuable resources for mental health research, providing a rich
source of data that can shed light on the prevalence, nature, and impact of these disorders. Researchers have utilized
social media data to assess mental health conditions, identify individuals at risk, and develop interventions to improve
well-being. The availability of massive datasets, coupled with advanced computational methods, has fueled the
development of novel approaches for mental health diagnosis using social media.

Mental illnesses, while traditionally diagnosed through clinical interviews and standardized assessments, are increasingly being detected through social media. With the widespread use of platforms such as X (formerly Twitter), Reddit, and Facebook, individuals often express their mental health struggles in online forums. This fact offers an unprecedented opportunity for researchers and clinicians to tap into large-scale, real-time data to identify patterns indicative of mental health conditions. Studies have shown that certain linguistic markers, such as negative sentiment, changes in language use, and self-referential statements, can be correlated with mental health issues like depression and anxiety\cite{De_Choudhury_De_2014,coppersmith-etal-2014-quantifying}.
For example, research has demonstrated that individuals with depression often use more first-person pronouns, indicating a heightened focus on the self, along with an increased use of negative emotion words such as "sad" or "hopeless". Similarly, anxiety-related posts frequently contain words reflecting uncertainty and worry, alongside expressions of physical symptoms like insomnia or fatigue. By mining these linguistic patterns, social media have become a valuable resource for detecting early signs of mental health deterioration\cite{GUNTUKU201743}.
However, there are also significant challenges in utilizing social media for mental illness diagnosis. Issues of privacy, the need for ethical guidelines, and the potential for misinterpretation of self-expressed symptoms are ongoing concerns\cite{resnik-etal-2015-beyond}. Despite these limitations, the potential for social media to provide real-time, large-scale insights into population-level mental health trends is a growing area of research, offering a complementary tool to traditional clinical diagnosis.

Recent advancements in natural language processing (NLP) and the development of large language models (LLMs) have significantly enhanced the potential for automated mental illness diagnosis. LLMs such as GPT, LLAMA and Phi, and their derivatives are capable of processing large volumes of unstructured text from social media and other platforms, allowing them to identify linguistic patterns associated with mental health conditions. These models can capture nuanced language use, which traditional machine learning methods may overlook.
For instance, LLMs have been used to detect signs of depression, anxiety, and even suicidal ideation through the analysis of social media posts. By leveraging pre-trained models, researchers have demonstrated that these models can outperform previous methods, such as lexicon-based approaches like LIWC and feature extraction techniques like TF-IDF, which rely on manual feature engineering and are limited in capturing nuanced linguistic patterns \cite{coppersmith-etal-2015-adhd,10.1145/3287560.3287587,matero-etal-2019-suicide}. These advancements highlight the ability of LLMs to capture subtle mental health signals, such as shifts in tone, sentiment, and specific word usage \cite{DBLP:journals/corr/abs-2005-14165}, providing a significant improvement over earlier approaches.

% \todo[inline]{DONE: methods like what, give examples and references.}

Furthermore, LLMs can be fine-tuned to adapt to the specific context of mental health data, improving their accuracy and making them highly versatile in handling diverse datasets.
% \todo[inline]{Add paragraph for LLMs explainability }
% One of the key advantages of LLMs in this domain is their ability to perform multi-label classification, where a single post can be labeled with multiple mental health conditions. This capability is crucial for detecting co-occurring mental illnesses, such as depression and anxiety, which are often diagnosed together in clinical settings \cite{10.1145/3673906}. Moreover, the scalability of LLMs enables large-scale deployment across various social media platforms, potentially providing real-time insights into mental health trends on a population level \cite{electronics12214396}.
One of the key advantages of LLMs in this domain is their ability to perform multi-label classification, where a single post can be labeled with multiple mental health conditions. This capability is crucial for detecting co-occurring mental illnesses, such as depression and anxiety, which are often diagnosed together in clinical settings \cite{10.1145/3673906}. Additionally, LLMs have the unique capability to not only provide classifications but also offer explanations for their decisions, making them particularly valuable in understanding the rationale behind a diagnosis \cite{zhao2023explainabilitylargelanguagemodels,luo2024understandingutilizationsurveyexplainability}. In the context of mental health, this can help clinicians better interpret why the model labeled a post with certain conditions based on linguistic features or patterns identified in the text. By providing these explanations, LLMs help bridge the gap between automated diagnosis and human understanding, ensuring that the outputs are interpretable and clinically relevant \cite{liu2024trustworthyllmssurveyguideline}. Moreover, the scalability of LLMs enables large-scale deployment across various social media platforms, potentially providing real-time insights into mental health trends on a population level \cite{electronics12214396}.

However, there are still challenges to overcome when applying LLMs to mental illness diagnosis. Ensuring model accuracy across diverse populations and linguistic variations, as well as addressing ethical concerns around privacy and the responsible use of data, remain important areas for ongoing research \cite{doi:10.1126/science.aal4230}. Despite these hurdles, LLMs represent a significant step forward in the automation and scalability of mental illness diagnosis, offering a promising tool for improving early detection and intervention efforts.
One of the key challenges in advancing automated mental illness diagnosis using large language models (LLMs) lies in the lack of publicly available multi-label mental illness datasets. Current datasets typically focus on single-label classification, where each instance (e.g., a social media post) is annotated for only one mental health condition, such as depression or anxiety. However, in real-world clinical settings, mental illnesses often co-occur; individuals may experience multiple conditions simultaneously, such as depression with anxiety or other mental disorders. The absence of multi-label datasets limits the ability of models to learn associations between these co-occurring conditions. Multi-label classification, where a post could be tagged with more than one condition, would enable more accurate and realistic diagnostic models. By identifying and addressing these associations, models could better reflect the complexity of mental health conditions and offer more nuanced insights for early intervention.

In the current research landscape, several studies explore LLMs for data annotation \cite{tan2024largelanguagemodelsdata} and synthetic labeling, predominantly focusing on enhancing data diversity and improving model performance for single-label classification tasks. However, these approaches do not address the need to expand single-label datasets into multi-label formats through zero-shot synthetic labeling—a technique that would allow LLMs to predict multiple potential labels for each instance without prior multi-label training \cite{yang2024datafreemultilabelimagerecognition,peskine-etal-2023-definitions}. This gap is particularly evident in mental health research, where multi-label datasets could drastically improve the diagnostic accuracy and effectiveness of models.
Despite the proven benefits of multi-label classification in other domains \cite{DBLP:journals/corr/abs-2102-07113}, mental health research has yet to see the same level of dataset development. This gap presents a significant barrier to progress, as the lack of comprehensive, multi-label datasets prevents models from fully capturing the intricacies of mental health symptoms. Without these datasets, LLMs cannot leverage the correlations between various mental illnesses, limiting their diagnostic accuracy and effectiveness.

This research focuses on leveraging the capabilities of large language models (LLMs) to address the critical gap in multi-label mental illness datasets. LLMs play a pivotal role in automating data annotation, offering a scalable and efficient solution for creating annotated datasets without requiring extensive manual effort. By using LLMs, more complex and nuanced datasets can be generated to capture the co-occurrence of mental health conditions. This automation allows for the rapid generation of labeled data, providing a valuable resource for future research and applications in mental health diagnosis.
The primary contribution of this work is the development of a method to synthetically transform any single-label dataset into a multi-label dataset using LLMs in a zero-shot setting. Unlike existing approaches that depend on labeled examples for each condition, this method enables LLMs to infer potential co-occurring conditions within a given text. This transformation converts single-label datasets into representations that capture a more realistic and multi-dimensional view of mental health.
Furthermore, this approach is applied to create a new, comprehensive multi-label mental illnesses dataset labeled for six distinct mental disorders. This dataset reflects co-occurring conditions and provides a valuable resource for training and evaluating multi-label classification models in mental health. This innovation not only simplifies the process of dataset annotation but also enhances data quality, making it better suited for real-world clinical applications. Ultimately, this approach improves the accuracy and relevance of predictive models for mental health diagnosis, paving the way for more advanced and comprehensive AI-driven mental health assessments. To the best of our knowledge, this is the first work to address this gap in the dataset and annotation.  

% \todo[inline]{DONE: we should mention the sections numbers; I fixed the first section, please do the same for other sections; label and cross-reference them }

The rest of this paper is organized as follows: Section \ref{sec:dataset_LLMs_metrics} provides a review of relevant datasets, including Dreaddit, DepSeverity, and the Reddit Mental Health Dataset (RMHD), as well as the large language models (LLMs) and evaluation metrics used. Section \ref{methodology} outlines the DepSeverity-Dreaddit multi-label dataset, describes the different prompt template types (single-label, multi-label, and unrestricted prompts), and details the evaluation process on multi-label datasets, including the SPAADE-DR dataset's preparation. Section \ref{results} presents the findings from evaluations on the DepSeverity-Dreaddit dataset and the SPAADE-DR dataset, covering aspects such as data cleaning, sampling, and multi-label labeling. Section \ref{analysis} explores the comorbidity between disorders and evaluates prompts and LLMs based on six key disorders. Finally, the paper concludes with a discussion in Section \ref{conclusion}, highlighting key takeaways and suggesting areas for future research.

% \todo[inline]{Done:We need a better title for this section, Related work does not represent what we have there.}
\section{Resources and Methods}
\label{sec:dataset_LLMs_metrics}
In this section, the key resources and methodologies underlying the study are presented, including the datasets, large language models (LLMs), and evaluation metrics used. By integrating multiple datasets and leveraging advanced LLMs, the goal is to develop a framework capable of capturing complex, real-world mental health patterns. The evaluation metrics are designed to support this goal, offering a robust framework to assess model performance in both single- and multi-label classifications.
\subsection{Datasets}
% In this paper we experimented using 6 main single label dataset and combined some of them together to create a new multi-label dataset and then developed a new method to label new .

% In this paper, we experimented with seven primary single-label datasets, each focusing on a specific mental health condition, along with the RMHD (Reddit Mental Health Dataset), which contains data on multiple mental health conditions. These datasets were individually labeled  for conditions such as depression, anxiety, stress, ADHD, eating disorders, PTSD, and suicide. To address the limitations of single-label datasets, we used them to create a new approach to label and convert datasets from single label to multi-label. We also created new multi-label dataset that captures the co-occurrence of multiple mental health conditions. This combination allowed us to develop and evaluate a novel method for annotating additional datasets, transforming them from single-label to multi-label datasets using large language models (LLMs) and synthetic labeling techniques. Below, we provide an overview of each dataset used in our research.

This paper experiments with seven primary single-label datasets, each focusing on a specific mental health condition, as well as the RMHD dataset, which contains data on multiple mental health conditions. These datasets are individually labeled for conditions like depression, anxiety, ADHD, eating disorders, PTSD, and emergent conditions, such as suicide. Below is an overview of each dataset used in this research.
% To overcome the limitations of single-label datasets, we developed an approach to convert these datasets into multi-label form, capturing the co-occurrence of multiple mental health conditions. Additionally, we created a new multi-label dataset that reflects the complexity of real-world mental health scenarios, where individuals may experience multiple conditions simultaneously.

% This combination enabled us to develop and evaluate a novel method for annotating other datasets, transforming them from single-label to multi-label using large language models (LLMs) and synthetic labeling techniques. 

\subsubsection{Dreaddit}
The Dreaddit dataset \cite{turcan2019dreaddit} contains 190,000 Reddit posts across ten subreddits in the five domains of abuse, social, anxiety, PTSD, and financial. These posts span a diverse range of content, including personal narratives, advice-seeking posts, and emotional expressions. Of these, 3,553 post segments are manually labeled with stress indicators by human annotators using Amazon Mechanical Turk, creating a supervised training set for stress detection. With its large scale, rich variety of content, and detailed annotations, Dreaddit is an invaluable resource for researchers studying stress in social media, providing insights across multiple real-world domains.

\subsubsection{DepSeverity}
The DepSeverity dataset \cite{naseem2022early} is derived from the same 3,553 Reddit posts used in the Dreaddit dataset, focusing on depression severity. These posts are labeled with four clinical levels of depression severity: Minimal, Mild, Moderate, and Severe, based on the Depressive Disorder Annotation (DDA) scheme\cite{mowery-etal-2015-towards}. The labeling process utilized six clinical resources: the Diagnostic and Statistical Manual of Mental Disorders, 5th Edition (DSM-5) \cite{APA2013}, the Behavioral Risk Factors Surveillance System (BRFSS), the Harvard Department of Psychiatry National Depression Screening Day Scale (HANDS), the Patient Health Questionnaire (PHQ-9)\cite{kroenke2002phq,kroenke2001phq}, the Quick Inventory of Depressive Symptomatology (QIDS-SR), and the Columbia Suicide Severity Rating Scale(C-SSRS) \cite{10.1145/3308558.3313698}.

Originally created as a binary dataset for detecting the presence of depression, it is later expanded to include four severity levels, providing a more nuanced understanding of users' mental health. This transformation helps better assess users’ condition, supporting potential intervention and treatment strategies. As noted by the dataset authors, at least 10 posts from a user are necessary for reliably predicting their depression severity, emphasizing the importance of longitudinal data for early intervention. With its detailed annotation and substantial volume of posts, DepSeverity offers a rich resource for researchers studying depression in social media. It spans various types of content such as personal narratives, advice-seeking, and emotional expressions, making it ideal for developing models that detect both the presence and severity of depression.

\subsubsection{Reddit Mental Health Dataset}The Reddit Mental Health Dataset (RMHD) \cite{covid2020} consists of a large collection of Reddit posts extracted from 28 subreddits, including both mental health-focused communities and general interest groups. The dataset spans posts from 17 mental health subreddits such as r/depression, r/Anxiety, and r/SuicideWatch, as well as 11 non-mental health subreddits such as r/conspiracy, r/legalagvice, r/personalfinance. 

The dataset includes posts from unique users across two critical timeframes: prepandemic (November 2018 to November 2019) and mid-pandemic (January to April 2020), allowing for comparative analyses of mental health trends before and during the COVID-19 pandemic. All posts are preprocessed to include only English-language content, and posts from bots, advertisements, and duplicate users are removed to ensure data quality. The subreddit each post belongs to is used as a label, associating the post with specific mental health conditions or general topics discussed in the community.

In addition to the raw text data, a variety of features are extracted from each post to enhance its utility for research. These include sentiment analysis using VADER, lexical counts, and the application of the Linguistic Inquiry and Word Count (LIWC) tool to assess semantic and grammatical categories such as emotions, pronouns, and body references. Moreover, lexicons are manually developed to track topics like suicidality, economic stress, isolation, and substance use, offering insights into the mental health conditions discussed in these posts. With its wide range of posts and comprehensive feature extraction, RMHD serves as a valuable resource for analyzing mental health trends, It is particularly suited for studies on mental health conditions, topic modeling, and trend analysis across large online communities.

\subsection{LLMs}
In this paper, experiments are done using five models of varying sizes, architectures, and model families to ensure comprehensive coverage across different approaches. These models are selected for their diversity in source and design, allowing for a robust comparison in performance.

\textbf{GPT-4o-mini}: Developed by OpenAI, GPT-4o-mini is a scaled-down variant of GPT-4o, designed to balance performance with computational efficiency. It features fewer parameters than the full GPT-4o model, making it more accessible for smaller-scale tasks without sacrificing core capabilities in natural language understanding and generation. The model is optimized for various NLP applications, such as text classification, summarization, and dialogue systems.

\textbf{Llama-3 70b}: Developed by Meta, Llama-3 70b is a 70-billion-parameter model designed to excel in a range of natural language processing tasks. It features multilingual support, coding capabilities, and advanced reasoning. Llama-3 benefits from an improved data curation pipeline and optimized scaling laws, allowing it to match the performance of leading models like GPT-4 across various benchmarks. Despite its size, Llama-3 has been fine-tuned to balance computational efficiency with high performance in both general and domain-specific tasks.

\textbf{Mistral NeMo 12b}: Mistral NeMo is a state-of-the-art model with 12 billion parameters, developed in collaboration with NVIDIA. It features an expansive 128k context window, offering exceptional performance in reasoning, world knowledge, and code generation within its size category. Mistral NeMo is particularly strong in multilingual applications, supporting a wide range of languages, including English, French, German, Chinese, and Arabic. The model uses the highly efficient Tekken tokenizer, which significantly improves compression rates across various languages and source code, outperforming previous Mistral models.

\textbf{Phi-3.5-MoE}: The Phi-3.5-MoE model, developed by Microsoft, uses a Mixture of Experts (MoE) architecture and features a total of 42 billion parameters, with 6.6 billion active parameters at any given time during inference. This model is designed to activate only a subset of its parameters (two out of sixteen experts) depending on the complexity of the task, which allows it to optimize resource usage while maintaining high performance across various natural language processing (NLP) tasks. This dynamic routing of experts enables the model to excel in diverse tasks such as reasoning, multilingual support, and code generation, while still being computationally efficient. By testing Phi-3 Medium, it is obvious that Phi-3.5-MoE performed better across our benchmarks, leading us to choose it over the Phi-3 Medium model.

\textbf{Gemma-2 9b}: Developed by Google DeepMind, Gemma-2 9b is a 9-billion-parameter model designed to deliver state-of-the-art performance in natural language understanding tasks. This model benefits from several enhancements, such as knowledge distillation and advanced attention mechanisms like Grouped-Query Attention (GQA). Despite its relatively smaller size compared to larger models, Gemma-2 9b demonstrates competitive performance across a range of benchmarks, offering a practical balance between size and accuracy for real-world applications.

\begin{table}[H]
\caption{Summary of large language models used in the experiments}
\label{tab:models}
\centering
{
% \resizebox{\textwidth}{!}{%
\begin{tabular}{@{}lcccc@{}}
\toprule
\multicolumn{1}{l}{\textbf{Model}} &
\multicolumn{1}{c}{\textbf{Size (Parameters)}} &
\multicolumn{1}{c}{\textbf{Source}} &
\multicolumn{1}{c}{\textbf{{API Service Provider} }} \\ \midrule

Gemma 2 9b \cite{team2024gemma2} & 9B & Google & Groq \\
GPT-4o mini \cite{openai2024gpt4omini} & Proprietary & OpenAI & OpenAI \\
Llama 3 70b \cite{dubey2024llama} & 70B & Meta & Groq \\
Mistral NeMo \cite{mistral2024nemo} & 7B & Mistral AI & Mistral AI \\ 
Phi-3.5-MoE \cite{abdin2024phi} & 42B & Phi-1 Labs & Azure AI \\     
\bottomrule
\end{tabular} \\
}

\end{table}

\subsection{Evaluation Metrics}
The evaluation of LLM performance is conducted using a variety of metrics, grouped into three categories: per-class, overall (multi-label), and multi-class. Metrics applied for per-class and multi-class evaluations include:
\begin{itemize}
    % \item Per category  Balanced Accuracy (CBA):
    % \item Per category  F1-Score (CF1): Binary F1-score
    % \item Per category  Precision (CP):
    % \item Per category  Recall (CR):
    % \item Overall Balanced Accuracy(OBA):
    % \item Overall F1-Score (OF1): Micro F1-Score
    % \item Overall Precision (OP):
    % \item Overall Recall (OR):
    % \item Hamming Loss (HL):
    % \item Multi-Class Balanced Accuracy (BA):
    
    \item \textbf{Balanced Accuracy (BA):}
    % \[
    % \text{BA} = \frac{\text{TPR} + \text{TNR}}{2}
    % \]
    \begin{equation}
    \text{\textbf{BA}} = \frac{1}{N} \sum_{i=0}^{N-1} \text{Recall}_i
    \label{eq:ba}
    \end{equation}
    where:
    % \[
    % \text{TPR} = \frac{\text{TP}}{\text{TP} + \text{FN}}, \quad \text{and} \quad   \text{TNR} = \frac{\text{TN}}{\text{TN} + \text{FP}}
    % \]
    \begin{itemize}
    \item \(N\): The total number of classes in the classification problem.
    \item \(\text{Recall}_i\): The recall for the \(i^\text{th}\) label or class.
\end{itemize}
    \item \textbf{F1-Score (F1):} (Binary F1-score)
    \[
    F1 = \frac{2 \cdot \text{Precision} \cdot \text{Recall}}{\text{Precision} + \text{Recall}}
    \]
    \item \textbf{Precision (CP):}
    \[
    \text{CP} = \frac{\text{TP}}{\text{TP} + \text{FP}}
    \]

    \item \textbf{Recall (CR):}
    \[
    \text{CR} = \frac{\text{TP}}{\text{TP} + \text{FN}}
    \]

\end{itemize}
where TP donates True Positives, FP False Positives, TN True Negatives, and FN False Negatives.

Balanced accuracy is chosen over regular accuracy to address the imbalance present in most datasets.
The F1-Score is incorporated to evaluate the model's performance, as it balances precision and recall, making it especially valuable when both false positives and false negatives have significant implications. For multi-label evaluation, the Micro F1-Score is employed to ensure equal consideration of all classes in the assessment.
Overall metrics, including Balanced Accuracy, Precision, and Recall, are calculated using the combined values of True Positives, False Positives, True Negatives, and False Negatives across all classes. These overall values are computed as follows:
\begin{itemize}
    \item \textbf{Overall TP, FP, TN, and FN:}
    \[
    \text{Overall TP} = \sum_{i=1}^{n} \text{TP}_i
    \]
    \[
    \text{Overall FP} = \sum_{i=1}^{n} \text{FP}_i
    \]
    \[
    \text{Overall TN} = \sum_{i=1}^{n} \text{TN}_i
    \]
    \[
    \text{Overall FN} = \sum_{i=1}^{n} \text{FN}_i
    \]
    where \( n \) is the number of classes or instances.

\end{itemize}
Hamming Loss is also utilized to evaluate performance in the multi-label classification setting. This metric quantifies the fraction of labels incorrectly predicted across all instances and labels, making it especially useful for multi-label tasks where each instance can have multiple associated labels. By focusing on individual label errors, Hamming Loss provides a granular view of the model’s accuracy across all labels, ensuring that each label’s misclassification is accounted for. It is calculated as follows:
\[
\text{Hamming Loss} = \frac{1}{N \cdot L} \sum_{i=1}^{N} \sum_{j=1}^{L} \mathbb{1}(y_{ij} \neq \hat{y}_{ij})
\]
where  \( N \) is the total number of instances, \( L \) is the total number of labels, \( y_{ij} \) is the true label for instance \( i \) and label \( j \), \( \hat{y}_{ij} \) is the predicted label for instance \( i \) and label \( j \), \( \mathbb{1}(\cdot) \) is the indicator function, which is 1 if \( y_{ij} \neq \hat{y}_{ij} \) and 0 otherwise.

A lower Hamming Loss indicates better model performance, with 0 representing perfect label prediction for every instance.

The Odds Ratio (OR), a measure of association between two events, is employed to analyze associations and comorbidities among mental disorders. Commonly used in medical and statistical analyses, the OR helps assess how strongly the presence or absence of one condition is associated with another. For two binary variables. This allows us to assess how likely one disorder is to occur alongside another, offering insights into potential comorbidities, the OR is calculated as follows:

\[
\text{OR} = \frac{\frac{a}{b}}{\frac{c}{d}} = \frac{a \times d}{b \times c}
\]

Where \textbf{a} = Number of cases where both the exposure and outcome are present (both event A and B occur). \textbf{b} = Number of cases where the exposure is present but the outcome is absent (event A occurs, but not event B). \textbf{c} = Number of cases where the exposure is absent but the outcome is present (event A does not occur, but event B occurs). \textbf{d} = Number of cases where neither the exposure nor the outcome is present (neither event A nor B occur).

\section{Methodology}
\label{methodology}

% \begin{tikzpicture}[node distance=1cm]
% % Nodes
% \node (start) [startstop] {Depseverity-Dreaddit Dataset\ref{dep-dre}};
% \node (single) [process, right of=start,] {Single-Label Prompt};
% \node (multi) [process, right of=start,yshift=-2cm] {Multi-Label Prompt};
% \node (unres) [process, right of=start,yshift=2cm] {Unrestricted Prompt};
% \node (eval) [process, right of=single] {LLMs and Prompts evaluation};
% \node (best) [decision, right of=eval] {Best Prompt and LLM};
% \node (comb) [process, below of=best] {Combined Dataset};
% \node (labeling) [process, left of=comb] {Semi-synthetic multi-label labeling};
% \node (labeled) [process, left of=labeling] {Multi-label Combined Dataset};
% \node (analysis) [startstop, left of=labeled] {Evaluation and Analysis};

% % Arrows
% \draw [arrow] (start) -- (single);
% \draw [arrow] (start) -- (multi);
% \draw [arrow] (start) -- (unres);
% \draw [arrow] (single) -- (eval);
% \draw [arrow] (multi) -- (eval);
% \draw [arrow] (unres) -- (eval);
% \draw [arrow] (eval) -- (best);
% \draw [arrow] (best) -- (labeling);
% \draw [arrow] (comb) -- (labeling);
% \draw [arrow] (labeling) -- (labeled);
% \draw [arrow] (labeled) -- (analysis);

% \end{tikzpicture}

\begin{figure}[ht]
    \centering
    \resizebox{\textwidth}{!}{%
        \begin{tikzpicture}[node distance=2.5cm and 3cm, scale=3]
            % Nodes
            \node(start) [sq] {Depseverity-Dreaddit Dataset (sec:\ref{dep-dre})};
            \node (single) [process, right of=start, xshift=1cm] {Single-Label Prompt (sec:\ref{single})};
            \node (multi) [process, below of=single, yshift=1cm] {Multi-Label Prompt (sec:\ref{multi})};
            \node (unres) [process, above of=single, yshift=-1cm] {Unrestricted Prompt (sec:\ref{unres})};
            \node (eval) [process2, right of=single, xshift=1.25cm] {LLMs and Prompts evaluation (sec:\ref{eval})};
            \node (best) [decision, right of=eval, xshift=1cm] {Best Prompt and LLM};
            \node (labeling) [process, below of=best, yshift=-0.5cm] {Semi-synthetic multi-label labeling};
            \node (comb) [process, right of=labeling, xshift=1.5cm] {SPAADE-DR Dataset};
            \node (labeled) [process, left of=labeling, xshift=-2cm] {Multi-label SPAADE-DR Dataset (sec:\ref{creation})};
            \node (analysis) [sq, left of=labeled, xshift=-1.5cm] {Evaluation and Analysis (sec:\ref{analysis})};

            % Arrows
            \draw [arrow] (start) -- (single);
            \draw [arrow] (start) -- (multi);
            \draw [arrow] (start) -- (unres);
            \draw [arrow] (single) -- (eval);
            \draw [arrow] (multi) -- (eval);
            \draw [arrow] (unres) -- (eval);
            \draw [arrow] (eval) -- (best);
            \draw [arrow] (best) -- (labeling);
            \draw [arrow] (comb) -- (labeling);
            \draw [arrow] (labeling) -- (labeled);
            \draw [arrow] (labeled) -- (analysis);
        \end{tikzpicture}
    }
        \caption{Process Workflow}
    \label{fig:workflow}
\end{figure}
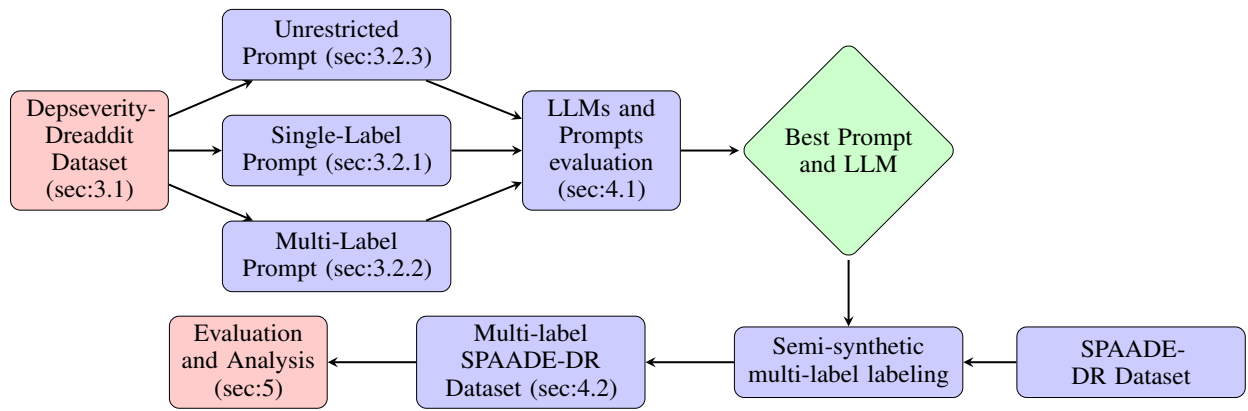

The methodology aims at developing a robust, multi-label dataset for mental health diagnostics by synthetically labeling social media posts, leveraging various prompt strategies with LLMs. This section outlines the structured workflow for dataset creation and labeling, as shown in Figure \ref{fig:workflow}. The Depseverity-Dreaddit merged dataset, labeled for depression and stress, is used to test three distinct prompt strategies—single-label, multi-label, and unrestricted—to guide LLMs in annotating multiple disorders. After evaluating the prompt effectiveness across LLMs, the optimal prompt-LLM combination is used to synthetically label RMHD dataset, resulting in a multi-label dataset that reflects complex mental health profiles.

\subsection{Depseverity-Dreaddit as Multi-Label Dataset}
\label{dep-dre}
The Depseverity and Dreaddit datasets consist of identical posts from the same user base, each independently labeled for either depression or stress. To utilize this overlap, these datasets are merged into a unified multi-label dataset, providing a resource labeled for both depression and stress. The resulting dataset distribution is shown in Table [\ref{tab...2.1}]. This new dataset, annotated by psychiatrists, enables a more nuanced analysis of mental health conditions, where users may exhibit symptoms of both disorders simultaneously. Merging these datasets enriches the available data for training and evaluating models capable of multi-label classification, thereby improving the accuracy and applicability of mental health detection tools.

\begin{table}[H]
\centering
\caption{The distribution of the depseverity-dreaddit dataset}
\label{tab...2.1}
\begin{tabular}{|c|ccc|}
\hline
Disorder & \multicolumn{3}{c|}{Stress}                                     \\ \hline
\multirow{3}{*}{Depression} & \multicolumn{1}{c|}{Label} & \multicolumn{1}{c|}{Negative} & Positive \\ \cline{2-4} 
         & \multicolumn{1}{c|}{Negative} & \multicolumn{1}{c|}{1532} & 1041 \\ \cline{2-4} 
         & \multicolumn{1}{c|}{Positive} & \multicolumn{1}{c|}{154}  & 814  \\ \hline
\end{tabular}
\end{table}

\subsection{Prompt Template}
The experiments utilize various types of prompts to diagnose mental illness in social media posts, with a particular focus on multi-label classification. Single-label, multi-label, and unrestricted prompts are tested across various datasets to evaluate their performance and efficiency. Each prompt template is paired with a specific parser to convert the model’s responses (e.g., "yes," "no," or the name of a disorder) into binary labels.
% \subsubsection{Single-Label Prompts}
% These prompts focused on diagnosing a single mental illness at a time. The model was prompted to assess whether the user showed symptoms of one specific disorder (e.g., depression or anxiety).
\subsubsection{Single-Label Prompts}
\label{single}
Single-label prompts are designed to diagnose one mental illness at a time, focusing the model’s attention on identifying whether the user exhibits symptoms of a specific disorder. This approach is simple yet effective for situations where only one condition is being evaluated. The model is prompted to make a binary decision, determining if the user shows clear signs of the targeted mental illness. A prompt inspired by those in \cite{hanafi2024comprehensiveevaluationlargelanguage} is utilized.

The following prompt template is used to guide the model in identifying a specific mental health condition:

\begin{figure}[ht]
\centering
\begin{tikzpicture}
\node[fill=gray!8, draw=gray, rounded corners, inner sep=5pt, text width=0.9\textwidth] (box) {
\begin{verbatim}
[Task]
Analyze the following social media post to determine if the writer exhibits clear 
symptoms of {The target disorder} according to provided guidelines.

[Guidelines]
- Concise Response: Respond only with 'Yes' (exhibits clear symptoms of {The target
  disorder}) or 'No' (Does not exhibit clear symptoms of {The target disorder}).
- No Explanations: Don't provide explanations for your assessment.
- Ambiguity: If the post is unclear, choose the most probable label.
[Post]
{The Post}
            \end{verbatim}
        };
    \end{tikzpicture}
        \caption{Single-label Binary Prompt Template for Identifying Mental Health Conditions}
    \label{fig:single_prompt_template}
\end{figure}

This prompt template in Fig:\ref{fig:single_prompt_template} allows for focused diagnosis of a single condition, minimizing confusion and maintaining clarity. The model’s binary response (Yes/No) delivers a straightforward mechanism for assessing the user’s symptoms according to specific clinical guidelines. When adapted for multi-label problems, this approach becomes resource-intensive, as it requires applying the same prompt multiple times—once for each mental illness or label being evaluated. While this method allows the model to focus exclusively on diagnosing a single illness per prompt, thereby reducing confusion, it incurs significantly higher computational costs.
\subsubsection{Multi-Label Prompts}\label{multi}

Various multi-label prompt templates are developed and tested to evaluate whether users exhibit multiple mental illnesses simultaneously. For this purpose, two primary approaches are identified:

\textbf{Template 1: Multi-Class Classification Approach}

This template frames the problem as a multi-class classification task. In cases with two mental illnesses (e.g., depression and stress), the model categorizes the post into one of four classes: "Depressed," "Stressed," "Depressed and Stressed," or "Normal". This approach accounts for scenarios where users display symptoms of more than one condition. However, as the number of mental illnesses (n) increases, the number of possible classes grows exponentially (2\textsuperscript{n}), which complicates classification and can confuse the model. As a result, this approach becomes less practical for higher numbers of labels due to the explosion in the number of classes.

\textbf{Template 2: Multi-Label Classification Approach}

This template treats the problem as a true multi-label classification task, asking the model to determine whether any of the n+1 conditions (including "Normal") is present. In the case of depression and stress, the model is prompted to assess whether the poster is "Depressed", "Stressed",  allowing for any combination of these labels or else he is "Normal". This method simplifies the classification task by avoiding the exponential class growth seen in Template 1. Instead of handling a growing number of classes, the model only deals with a manageable set of label combinations, making it more efficient as the number of conditions increases.

The multi-label approach ensures greater flexibility in identifying overlapping symptoms and prevents the complexity associated with multi-class classification. Both templates are evaluated for their effectiveness in addressing posts that exhibit co-occurring mental health symptoms.
The prompt templates in Fig: \ref{fig:multi_label_templates} guides the model in identifying a number of specific mental health conditions, in this case depression and stress.
\begin{figure}[ht]
\centering
\begin{tikzpicture}
\node[fill=gray!8, draw=gray, rounded corners, inner sep=5pt, text width=0.92\textwidth] (box) {
\begin{verbatim}
[Task]
Analyze the following social media post to determine if the writer exhibits clear 
symptoms of {Depression or Stress} according to provided guidelines.

[Guidelines]
- The poster could have multiple illnesses at the same time; otherwise, he is normal.
- Concise Response: Respond with one of these 4 words only ["Depressed", "Stressed",
  "Depressed and Stressed", "Normal"].
- No Explanations: Don't provide explanations for your assessment.
- Ambiguity: If the post is unclear, choose the most probable label.
[Post]
{The Post}
            \end{verbatim}
        };
\end{tikzpicture}

\textbf{Multi Label Binary Template 1}

\vspace{0.5cm}
\begin{tikzpicture}
\node[fill=gray!8, draw=gray, rounded corners, inner sep=5pt, text width=0.92\textwidth] (box) {
\begin{verbatim}
[Task]
As a psychiatrist, analyze the provided social media post to determine if the writer 
exhibits clear symptoms of {Depression or Stress} according to provided guidelines.

[Guidelines]
- The poster could have multiple illnesses at the same time; otherwise, he is normal.
- Concise Response: Respond with any combination of these 2 mental illness names or
  "Normal" only ["Depressed", "Stressed"]. If the poster doesn't have any 
  mental illness, just answer with "Normal".
- No Explanations: Don't provide explanations for your assessment.
- Ambiguity: If the post is unclear, choose the most probable label.
[Post]
{The post}
            \end{verbatim}
        };
    \end{tikzpicture}
    
\textbf{Multi Label Binary Template 2}

        \caption{Multi-Label Binary Prompt Templates for Identifying Mental Health Conditions}
    \label{fig:multi_label_templates}
\end{figure}
These templates are designed to provide clear and concise assessments while maintaining the flexibility to handle multiple conditions, making them effective tools for multi-label classification tasks.
% \subsubsection{Unrestricted Prompt}
% We also experimented with unrestricted prompts to diagnose multiple mental illnesses without limiting the model to predefined categories. The goal of this approach was to provide the model with the freedom to detect any mental health condition based on the content of the post.

% These unrestricted prompts were assessed to gauge their effectiveness in identifying a broader range of mental health issues and compared against the more structured single-label and multi-label prompt types.
\subsubsection{Unrestricted Prompt}
\label{unres}
Unrestricted prompts are also explored, enabling the model to diagnose multiple mental illnesses without being restricted to predefined categories. The aim of this approach is to give the model more freedom in detecting any mental health condition based solely on the content of the post, rather than being confined to a limited set of labels. The unrestricted prompt is designed to assess the model's ability to identify a broader spectrum of mental health issues and are compared to the more structured single-label and multi-label prompts. The flexibility of this approach enables the model to diagnose multiple co-occurring conditions or determine if the poster exhibits no symptoms at all.

\begin{figure}[H]
    
    \centering
    \begin{tikzpicture}
        \node[fill=gray!8, draw=gray, rounded corners, inner sep=5pt, text width=0.98\textwidth] (box) {
        \begin{verbatim}
As a psychiatrist, analyze the provided social media post to determine if the writer 
exhibits clear symptoms of any mental illness according to provided guidelines.

[Guidelines]
- The poster could have multiple illnesses at the same time; otherwise, he is normal.
- Without explaining your reasoning, respond with any combination of mental illnesses
  names separated by comas only and if the poster doesn't have any mental ilness just 
  answer with "Normal"
- Answer with basic mental illnesses names only like "Depression" without further details or 
  complex names.
[Post]
{The post}
        \end{verbatim}
        };
    \end{tikzpicture}
    \caption{Unrestricted Binary Prompt Template for Identifying Mental Health Conditions}
    \label{fig:unres-template}
\end{figure}

This template in Fig:\ref{fig:unres-template} provides the model with the flexibility to identify a wide range of mental health conditions without the constraints of predefined labels, making it a valuable tool for exploring diverse mental health issues in user-generated content.

\subsection{Evaluation on a Multi-label dataset}

The performance of various prompt templates and models is evaluated using the Depseverity-Dreaddit dataset. The goal is to identify the most effective prompt-template and model combinations for accurate multi-label classification. 
Additionally, the impact of majority voting across model predictions is examined to determine its potential for improving accuracy and reducing biases in the results. This analysis provides insights into how different prompts and models handle overlapping mental health conditions in a multi-label context.
\subsection{SPAADE-DR dataset}
% After an extensive search, we found no existing accurate multi-label datasets for mental health classification, so we decided to create a new semi-synthetic multi-labeled dataset. This was achieved by combining multiple datasets, each focused on a single mental health condition.We utilized the RMHD dataset for this task because it contains datasets for multiple mental disorders, we used the datasets for the disorders ADHD, anxiety,depression, eating disorder, PTSD, suicide. By merging these single-label datasets, we created a comprehensive dataset that covers multiple labels, allowing for more complex, multi-label labeling, analysis and model evaluation.
An extensive search reveals a lack of existing, accurate multi-label datasets for mental health classification. To fill this gap, a new semi-synthetic multi-label dataset, SPAADE-DR (Suicidal Ideation, PTSD, Anxiety, ADHD, Depression, Eating Disorder Diagnosis from Reddit posts), is created by combining and labeling multiple single-label datasets, each dedicated to a specific mental health condition. For this task, the RMHD dataset is utilized, encompassing data for various mental disorders. Specifically, datasets for conditions such as ADHD, anxiety, depression, eating disorders, PTSD, and suicide are incorporated. By merging these single-label datasets, a comprehensive multi-class dataset is constructed to represent multiple mental health conditions. This allows for more sophisticated multi-label labeling, analysis, and model evaluation, providing a valuable resource for further research into mental health classification. See section:\ref{creation} for more details.

Following the evaluation of the prompt templates and model performance, the optimal combinations are applied to label the SPAADE-DR dataset. 
% This dataset, which integrates six single-label datasets (covering ADHD, anxiety, depression, eating disorders, PTSD,and suicide), was originally annotated for one mental health condition per post.
the best-performing prompt-template and model combinations identified from the Depseverity-Dreaddit evaluation are utilized to predict and label the remaining five conditions for each post, ensuring consistency and accuracy in the multi-label classification.
After labeling the SPAADE-DR dataset, it is used to evaluate the performance of both multi-label and unrestricted prompt templates on the newly created six-label dataset. This evaluation aims to assess how well the models perform when the number of labels increases in a multi-class setting. Additionally, it allows us to explore the models' ability to identify correlations between various mental illnesses and their associated symptoms. Testing both multi-label and unrestricted prompts aims to uncover the strengths and limitations of the models in capturing the complexity of overlapping mental health conditions.

\section{Results \& Discussion}
\label{results}
This section presents the results of the synthetic labeling methodology, highlighting the effectiveness of various prompt strategies and their influence on multi-label mental health diagnostics. Each prompt-LLM combination is evaluated on the Depseverity-Dreaddit dataset to analyze how single-label, multi-label, and unrestricted prompts affect diagnostic accuracy across various disorders. Insights gained from this analysis inform the selection of the optimal prompt-LLM configuration for creating a nuanced multi-label dataset. Finally, the implications of the results are discussed, emphasizing the dataset's potential to advance multi-disorder mental health assessment in social media contexts.

\subsection{Evaluation on Depseverity-Dreaddit Dataset}
\label{eval}
In the initial experiment, various LLMs and prompt templates are evaluated using a balanced subset of the Depseverity-Dreaddit dataset. The goal of this evaluation is to identify the most effective LLM-prompt combinations for accurate multi-label classification of mental health conditions. Additionally, majority voting across models is tested to assess its impact on improving overall accuracy.

The metrics chosen are used to assess the models’ ability to predict each label (depression and stress) separately, their overall multi-label classification accuracy, and their performance in multi-class classification (combining all possible classes). From the results in Table \ref{tab..exp1}, it is clear that the best-performing LLM is Llama-3 70b, closely followed by GPT-4o-mini and Phi-3.5 MoE. Llama-3 70b demonstrates the highest scores using all prompts across most metrics, particularly in multi-label classification, where it achieves the highest overall balanced accuracy (0.78) with a lower hamming loss (0.24). GPT-4o-mini and Phi-3.5 MoE also perform strongly, with competitive results in the same metrics, making them viable alternatives.

In terms of prompt templates, the single-label prompt template consistently outperforms the multi-label templates. This is evident from the higher precision, recall, and F1-scores when evaluating the LLMs on each label (depression and stress). The single-label prompts provide a more focused diagnosis for each condition, contributing to the models’ higher accuracy in both multi-label and multi-class evaluations. Although multi-label templates capture co-occurring conditions, they generally result in lower precision and higher Hamming loss, which indicates some confusion in classifying posts with overlapping symptoms.

Additionally, after testing majority voting between models to see if combining predictions from multiple LLMs improves accuracy, it is obvious that while majority voting slightly improves recall, it does not significantly impact precision or F1-scores, and in some cases, it leads to higher Hamming loss. As a result, majority voting does not offer a clear advantage over individual model predictions. In conclusion, for multi-label classification on the Depseverity-Dreaddit dataset, the combination of Llama-3 70b and the single-label prompt template delivers the most accurate and reliable results. This combination effectively handles the complexities of diagnosing both depression and stress simultaneously.

\begin{landscape}
\centering
\begin{table}[h]
\centering
\caption{Comparisons between prompt templates and LLMs on the Depseverity-Dreaddit dataset are conducted using several metrics: per-category balanced accuracy (CBA), F1-measure (CF1), precision (CP), recall (CR), overall balanced accuracy (OBA), overall precision (OP), and overall recall (OR)\cite{DBLP:journals/corr/abs-2107-10834,yao2024gkgnetgroupknearestneighbor}, hamming loss (HL)\cite{DBLP:journals/corr/abs-2011-07805}, and Multi-class balanced accuracy (BA).}
\label{tab..exp1}
% \resizebox{\textwidth}{!}
{%
\begin{tabular}{c|c||cccc||cccc||ccccc||c}
\hline\hline
\multirow{2}{*}{Prompt} &
  \multirow{2}{*}{LLM} &
  \multicolumn{4}{c||}{Depression} &
  \multicolumn{4}{c||}{Stress} &
  \multicolumn{5}{c||}{Multi-Label} &
  Multi-Class \\ \cline{3-15}
 &
   &
  \multicolumn{1}{c|}{CBA} &
  \multicolumn{1}{c|}{CF1} &
  \multicolumn{1}{c|}{CP} &
  CR &
  \multicolumn{1}{c|}{CBA} &
  \multicolumn{1}{c|}{CF1} &
  \multicolumn{1}{c|}{CP} &
  CR &
  \multicolumn{1}{c|}{GBA} &
  \multicolumn{1}{c|}{OF1} &
  \multicolumn{1}{c|}{OP} &
  \multicolumn{1}{c|}{OR} &
  HL &
  BA \\ \hline\hline
\multirow{6}{*}{Single-Label} &
  Lama-3 70b &
  \multicolumn{1}{c|}{0.74} &
  \multicolumn{1}{c|}{0.61} &
  \multicolumn{1}{c|}{\textbf{0.54}} &
  0.71 &
  \multicolumn{1}{c|}{\textbf{0.75}} &
  \multicolumn{1}{c|}{\textbf{0.81}} &
  \multicolumn{1}{c|}{\textbf{0.70}} &
  0.96 &
  \multicolumn{1}{c|}{\underline{\textbf{0.78}}} &
  \multicolumn{1}{c|}{\underline{\textbf{0.74}}} &
  \multicolumn{1}{c|}{\underline{\textbf{0.64}}} &
  \multicolumn{1}{c|}{0.88} &
  \underline{\textbf{0.24}} &
  \underline{\textbf{0.46}} \\
 &
  Gemma-2 9b &
  \multicolumn{1}{c|}{0.72} &
  \multicolumn{1}{c|}{0.58} &
  \multicolumn{1}{c|}{0.44} &
  0.84 &
  \multicolumn{1}{c|}{0.64} &
  \multicolumn{1}{c|}{0.75} &
  \multicolumn{1}{c|}{0.60} &
  \underline{\textbf{1.00}} &
  \multicolumn{1}{c|}{0.71} &
  \multicolumn{1}{c|}{0.69} &
  \multicolumn{1}{c|}{0.54} &
  \multicolumn{1}{c|}{0.94} &
  0.34 &
  0.39 \\
 &
  Phi-3.5-MoE &
  \multicolumn{1}{c|}{\underline{\textbf{0.75}}} &
  \multicolumn{1}{c|}{\underline{\textbf{0.62}}} &
  \multicolumn{1}{c|}{0.52} &
  0.75 &
  \multicolumn{1}{c|}{0.74} &
  \multicolumn{1}{c|}{0.79} &
  \multicolumn{1}{c|}{\textbf{0.70}} &
  0.92 &
  \multicolumn{1}{c|}{0.77} &
  \multicolumn{1}{c|}{0.73} &
  \multicolumn{1}{c|}{0.63} &
  \multicolumn{1}{c|}{0.87} &
  0.25 &
  \underline{\textbf{0.46}} \\
 &
  GPT-4o-mini &
  \multicolumn{1}{c|}{0.74} &
  \multicolumn{1}{c|}{0.61} &
  \multicolumn{1}{c|}{0.47} &
  0.84 &
  \multicolumn{1}{c|}{0.65} &
  \multicolumn{1}{c|}{0.76} &
  \multicolumn{1}{c|}{0.61} &
  \underline{\textbf{1.00}} &
  \multicolumn{1}{c|}{0.73} &
  \multicolumn{1}{c|}{0.71} &
  \multicolumn{1}{c|}{0.56} &
  \multicolumn{1}{c|}{0.94} &
  0.31 &
  0.40 \\
 &
  Mistral-Nemo &
  \multicolumn{1}{c|}{0.59} &
  \multicolumn{1}{c|}{0.48} &
  \multicolumn{1}{c|}{0.32} &
  \underline{\textbf{0.96}} &
  \multicolumn{1}{c|}{0.52} &
  \multicolumn{1}{c|}{0.70} &
  \multicolumn{1}{c|}{0.54} &
  \underline{\textbf{1.00}} &
  \multicolumn{1}{c|}{0.57} &
  \multicolumn{1}{c|}{0.60} &
  \multicolumn{1}{c|}{0.44} &
  \multicolumn{1}{c|}{\underline{\textbf{0.99}}} &
  0.51 &
  0.27 \\
 &
  Majority Vote &
  \multicolumn{1}{c|}{0.72} &
  \multicolumn{1}{c|}{0.58} &
  \multicolumn{1}{c|}{0.43} &
  0.88 &
  \multicolumn{1}{c|}{0.63} &
  \multicolumn{1}{c|}{0.75} &
  \multicolumn{1}{c|}{0.60} &
  \underline{\textbf{1.00}} &
  \multicolumn{1}{c|}{0.70} &
  \multicolumn{1}{c|}{0.69} &
  \multicolumn{1}{c|}{0.53} &
  \multicolumn{1}{c|}{0.95} &
  0.35 &
  0.38 \\ \hline
\multirow{6}{*}{Multi-Label-1} &
  Llama-3 70b &
  \multicolumn{1}{c|}{0.71} &
  \multicolumn{1}{c|}{0.58} &
  \multicolumn{1}{c|}{0.47} &
  0.73 &
  \multicolumn{1}{c|}{\underline{\textbf{0.78}}} &
  \multicolumn{1}{c|}{\underline{\textbf{0.82}}} &
  \multicolumn{1}{c|}{\textbf{0.73}} &
  0.93 &
  \multicolumn{1}{c|}{\textbf{0.76}} &
  \multicolumn{1}{c|}{\textbf{0.73}} &
  \multicolumn{1}{c|}{\textbf{0.63}} &
  \multicolumn{1}{c|}{0.87} &
  \textbf{0.26} &
  \textbf{0.48} \\
 &
  Gemma-2 9b &
  \multicolumn{1}{c|}{0.69} &
  \multicolumn{1}{c|}{0.55} &
  \multicolumn{1}{c|}{0.46} &
  0.69 &
  \multicolumn{1}{c|}{0.64} &
  \multicolumn{1}{c|}{0.75} &
  \multicolumn{1}{c|}{0.61} &
  0.98 &
  \multicolumn{1}{c|}{0.71} &
  \multicolumn{1}{c|}{0.68} &
  \multicolumn{1}{c|}{0.56} &
  \multicolumn{1}{c|}{0.88} &
  0.32 &
  0.39 \\
 &
  Phi-3.5-MoE &
  \multicolumn{1}{c|}{\textbf{0.73}} &
  \multicolumn{1}{c|}{\textbf{0.59}} &
  \multicolumn{1}{c|}{0.47} &
  \textbf{0.79} &
  \multicolumn{1}{c|}{0.66} &
  \multicolumn{1}{c|}{0.74} &
  \multicolumn{1}{c|}{0.63} &
  0.89 &
  \multicolumn{1}{c|}{0.71} &
  \multicolumn{1}{c|}{0.68} &
  \multicolumn{1}{c|}{0.57} &
  \multicolumn{1}{c|}{0.86} &
  0.32 &
  0.42 \\
 &
  GPT-4o-mini &
  \multicolumn{1}{c|}{0.71} &
  \multicolumn{1}{c|}{0.57} &
  \multicolumn{1}{c|}{0.45} &
  0.77 &
  \multicolumn{1}{c|}{0.72} &
  \multicolumn{1}{c|}{0.79} &
  \multicolumn{1}{c|}{0.67} &
  0.96 &
  \multicolumn{1}{c|}{0.74} &
  \multicolumn{1}{c|}{0.71} &
  \multicolumn{1}{c|}{0.59} &
  \multicolumn{1}{c|}{\textbf{0.89}} &
  0.29 &
  0.44 \\
 &
  Mistral-Nemo &
  \multicolumn{1}{c|}{0.64} &
  \multicolumn{1}{c|}{0.45} &
  \multicolumn{1}{c|}{\underline{\textbf{0.65}}} &
  0.34 &
  \multicolumn{1}{c|}{0.59} &
  \multicolumn{1}{c|}{0.71} &
  \multicolumn{1}{c|}{0.58} &
  0.93 &
  \multicolumn{1}{c|}{0.70} &
  \multicolumn{1}{c|}{0.65} &
  \multicolumn{1}{c|}{0.59} &
  \multicolumn{1}{c|}{0.73} &
  0.31 &
  0.36 \\
 &
  Majority Vote &
  \multicolumn{1}{c|}{0.71} &
  \multicolumn{1}{c|}{0.58} &
  \multicolumn{1}{c|}{0.49} &
  0.71 &
  \multicolumn{1}{c|}{0.66} &
  \multicolumn{1}{c|}{0.76} &
  \multicolumn{1}{c|}{0.62} &
  \textbf{0.99} &
  \multicolumn{1}{c|}{0.73} &
  \multicolumn{1}{c|}{0.70} &
  \multicolumn{1}{c|}{0.58} &
  \multicolumn{1}{c|}{\textbf{0.89}} &
  0.30 &
  0.42 \\ \hline
\multirow{6}{*}{Multi-Label-2} &
  Llama-3 70b &
  \multicolumn{1}{c|}{\textbf{0.74}} &
  \multicolumn{1}{c|}{\textbf{0.61}} &
  \multicolumn{1}{c|}{0.54} &
  0.70 &
  \multicolumn{1}{c|}{\textbf{0.71}} &
  \multicolumn{1}{c|}{\textbf{0.76}} &
  \multicolumn{1}{c|}{\textbf{0.69}} &
  0.85 &
  \multicolumn{1}{c|}{\textbf{0.75}} &
  \multicolumn{1}{c|}{\textbf{0.71}} &
  \multicolumn{1}{c|}{\textbf{0.64}} &
  \multicolumn{1}{c|}{0.80} &
  \textbf{0.26} &
  \textbf{0.48} \\
 &
  Gemma-2 9b &
  \multicolumn{1}{c|}{0.70} &
  \multicolumn{1}{c|}{0.55} &
  \multicolumn{1}{c|}{0.41} &
  0.84 &
  \multicolumn{1}{c|}{0.63} &
  \multicolumn{1}{c|}{0.75} &
  \multicolumn{1}{c|}{0.60} &
  0.98 &
  \multicolumn{1}{c|}{0.69} &
  \multicolumn{1}{c|}{0.67} &
  \multicolumn{1}{c|}{0.53} &
  \multicolumn{1}{c|}{0.93} &
  0.36 &
  0.37 \\
 &
  Phi-3.5-MoE &
  \multicolumn{1}{c|}{0.72} &
  \multicolumn{1}{c|}{0.58} &
  \multicolumn{1}{c|}{0.45} &
  0.82 &
  \multicolumn{1}{c|}{0.67} &
  \multicolumn{1}{c|}{0.72} &
  \multicolumn{1}{c|}{0.65} &
  0.79 &
  \multicolumn{1}{c|}{0.70} &
  \multicolumn{1}{c|}{0.66} &
  \multicolumn{1}{c|}{0.57} &
  \multicolumn{1}{c|}{0.80} &
  0.32 &
  0.42 \\
 &
  GPT-4o-mini &
  \multicolumn{1}{c|}{0.69} &
  \multicolumn{1}{c|}{0.56} &
  \multicolumn{1}{c|}{\textbf{0.60}} &
  0.52 &
  \multicolumn{1}{c|}{0.66} &
  \multicolumn{1}{c|}{0.72} &
  \multicolumn{1}{c|}{0.64} &
  0.83 &
  \multicolumn{1}{c|}{0.72} &
  \multicolumn{1}{c|}{0.67} &
  \multicolumn{1}{c|}{0.63} &
  \multicolumn{1}{c|}{0.72} &
  0.28 &
  0.44 \\
 &
  Mistral-Nemo &
  \multicolumn{1}{c|}{0.70} &
  \multicolumn{1}{c|}{0.56} &
  \multicolumn{1}{c|}{0.41} &
  \textbf{0.86} &
  \multicolumn{1}{c|}{0.53} &
  \multicolumn{1}{c|}{0.70} &
  \multicolumn{1}{c|}{0.54} &
  \underline{\textbf{1.00}} &
  \multicolumn{1}{c|}{0.65} &
  \multicolumn{1}{c|}{0.65} &
  \multicolumn{1}{c|}{0.49} &
  \multicolumn{1}{c|}{\textbf{0.95}} &
  0.41 &
  0.32 \\
 &
  Majority Vote &
  \multicolumn{1}{c|}{\textbf{0.74}} &
  \multicolumn{1}{c|}{0.60} &
  \multicolumn{1}{c|}{0.50} &
  0.75 &
  \multicolumn{1}{c|}{0.64} &
  \multicolumn{1}{c|}{0.75} &
  \multicolumn{1}{c|}{0.61} &
  0.96 &
  \multicolumn{1}{c|}{0.73} &
  \multicolumn{1}{c|}{0.70} &
  \multicolumn{1}{c|}{0.57} &
  \multicolumn{1}{c|}{0.89} &
  0.31 &
  0.42 \\ \hline
\multirow{6}{*}{Unrestricted} &
  Llama-3 70b &
  \multicolumn{1}{c|}{\textbf{0.71}} &
  \multicolumn{1}{c|}{\textbf{0.57}} &
  \multicolumn{1}{c|}{\textbf{0.48}} &
  0.69 &
  \multicolumn{1}{c|}{0.71} &
  \multicolumn{1}{c|}{0.76} &
  \multicolumn{1}{c|}{0.69} &
  \textbf{0.85} &
  \multicolumn{1}{c|}{\textbf{0.73}} &
  \multicolumn{1}{c|}{\textbf{0.69}} &
  \multicolumn{1}{c|}{\textbf{0.61}} &
  \multicolumn{1}{c|}{0.80} &
  \textbf{0.28} &
  \textbf{0.44} \\
 &
  Gemma-2 9b &
  \multicolumn{1}{c|}{0.70} &
  \multicolumn{1}{c|}{0.56} &
  \multicolumn{1}{c|}{0.42} &
  0.84 &
  \multicolumn{1}{c|}{0.72} &
  \multicolumn{1}{c|}{0.74} &
  \multicolumn{1}{c|}{0.72} &
  0.76 &
  \multicolumn{1}{c|}{0.70} &
  \multicolumn{1}{c|}{0.66} &
  \multicolumn{1}{c|}{0.57} &
  \multicolumn{1}{c|}{0.79} &
  0.32 &
  0.43 \\
 &
  Phi-3.5-MoE &
  \multicolumn{1}{c|}{0.70} &
  \multicolumn{1}{c|}{0.55} &
  \multicolumn{1}{c|}{0.41} &
  0.85 &
  \multicolumn{1}{c|}{\textbf{0.73}} &
  \multicolumn{1}{c|}{0.74} &
  \multicolumn{1}{c|}{\underline{\textbf{0.74}}} &
  0.74 &
  \multicolumn{1}{c|}{0.69} &
  \multicolumn{1}{c|}{0.66} &
  \multicolumn{1}{c|}{0.57} &
  \multicolumn{1}{c|}{0.78} &
  0.32 &
  0.42 \\
 &
  GPT-4o-mini &
  \multicolumn{1}{c|}{0.68} &
  \multicolumn{1}{c|}{0.54} &
  \multicolumn{1}{c|}{0.40} &
  0.80 &
  \multicolumn{1}{c|}{\textbf{0.73}} &
  \multicolumn{1}{c|}{\textbf{0.77}} &
  \multicolumn{1}{c|}{0.72} &
  0.82 &
  \multicolumn{1}{c|}{0.70} &
  \multicolumn{1}{c|}{0.67} &
  \multicolumn{1}{c|}{0.57} &
  \multicolumn{1}{c|}{\textbf{0.82}} &
  0.32 &
  0.41 \\
 &
  Mistral-Nemo &
  \multicolumn{1}{c|}{0.51} &
  \multicolumn{1}{c|}{0.43} &
  \multicolumn{1}{c|}{0.28} &
  \textbf{0.91} &
  \multicolumn{1}{c|}{0.52} &
  \multicolumn{1}{c|}{0.63} &
  \multicolumn{1}{c|}{0.54} &
  0.78 &
  \multicolumn{1}{c|}{0.50} &
  \multicolumn{1}{c|}{0.54} &
  \multicolumn{1}{c|}{0.40} &
  \multicolumn{1}{c|}{\textbf{0.82}} &
  0.57 &
  0.27 \\
 &
  Majority Vote &
  \multicolumn{1}{c|}{0.68} &
  \multicolumn{1}{c|}{0.54} &
  \multicolumn{1}{c|}{0.41} &
  0.79 &
  \multicolumn{1}{c|}{0.72} &
  \multicolumn{1}{c|}{0.76} &
  \multicolumn{1}{c|}{0.71} &
  0.81 &
  \multicolumn{1}{c|}{0.70} &
  \multicolumn{1}{c|}{0.67} &
  \multicolumn{1}{c|}{0.57} &
  \multicolumn{1}{c|}{0.81} &
  0.32 &
  0.42 \\ \hline\hline
\end{tabular}%
}
\end{table}
\end{landscape}

\subsection{SPAADE-DR Dataset}

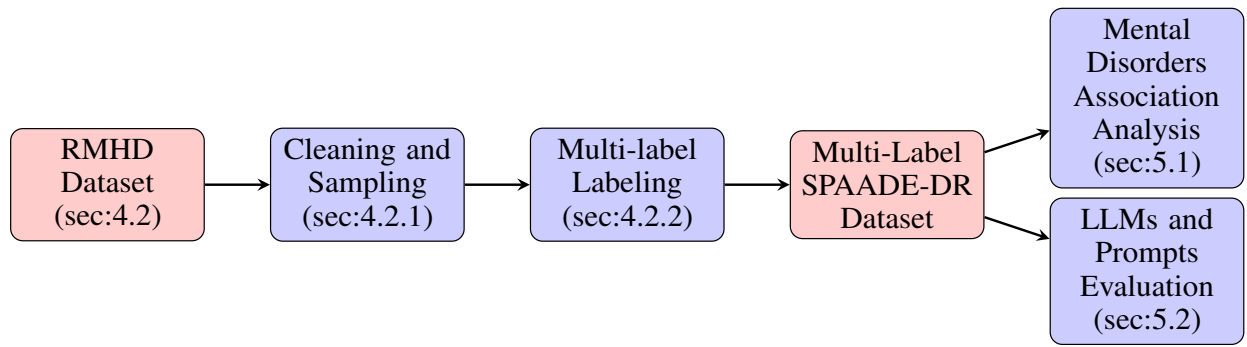
\begin{figure}[H]
    \centering
    \resizebox{\textwidth}{!}{%
        \begin{tikzpicture}[node distance=2.5cm and 3cm, scale=3]
            % Nodes
            \node(crt) [sq] {RMHD Dataset (sec:\ref{creation})};
            \node(c_s) [process2, right of=crt, xshift=0.5cm] {Cleaning and Sampling (sec:\ref{cleaning_and_sampling})};
            \node(la) [process2, right of=c_s, xshift=0.5cm] {Multi-label Labeling (sec:\ref{labeling_})};
            \node(ds) [sq, right of=la, xshift=0.5cm] {Multi-Label SPAADE-DR Dataset};
            \node(anl) [process2, right of=ds, xshift=0.5cm, yshift=1cm] {Mental Disorders Association Analysis (sec:\ref{analysis2})};
            \node(eva) [process2, right of=ds, xshift=0.5cm, yshift=-1cm] {LLMs and Prompts Evaluation (sec:\ref{reeveal})};

            % Arrows
            \draw [arrow] (crt) -- (c_s);
            \draw [arrow] (c_s) -- (la);
            \draw [arrow] (la) -- (ds);
            \draw [arrow] (ds) -- (anl);
            \draw [arrow] (ds) -- (eva);
        \end{tikzpicture}
    }
    \caption{Workflow for the SPAADE-DR Dataset Process}
    \label{fig:workflow_rmhd}
\end{figure}

\label{creation}
As mentioned earlier, the RMHD dataset originally consists of posts from 17 mental health subreddits and 11 non-mental health subreddits. For this study, posts from r/conspiracy, r/jokes, r/teaching, r/personalfinance, and r/legaladvice are selected as control samples, and posts from r/adhd, r/anxiety, r/depression, r/EDAnonymous (Eating Disorder), r/ptsd, and r/suicidewatch as samples representing various mental health disorders. Since the dataset uses the subreddit each post originates from as its label, it is not entirely clean. Some posts are incorrectly labeled as positive for a disorder simply because they are posted in a particular mental health subreddit, even if they do not reflect the condition.To address this, a cleaning and sampling process is applied, followed by multi-label labeling, as outlined in the workflow shown in Figure \ref{fig:workflow_rmhd}. 

\subsubsection{Cleaning and Sampling}
\label{cleaning_and_sampling}
Initially, the dataset is cleaned using unsupervised and semi-supervised learning techniques, including KMeans, DBSCAN, and One-Class SVM. These methods leverage both the features provided with the dataset and word embeddings to identify and remove outliers or irrelevant posts. However, these techniques fail to produce satisfactory results. Furthermore, due to the large size of the RMHD dataset, a smaller, cleaner subset must be sampled for further analysis.

To address both cleaning and sampling efficiently, 600 posts are initially sampled from each mental health disorder subreddit. The LLaMA-3 70b model, identified as the best-performing LLM, is then employed to evaluate and classify these posts. Based on the model's predictions, posts predicted as negative for the corresponding disorder are manually reviewed, and true negatives are removed. Following the cleaning process, 500 posts are selected from the remaining subset of 600 for each disorder, ensuring a clean and representative sample for further analysis.

\subsubsection{Multi-label labeling}
\label{labeling_}
% \todo[inline]{which prompt?? DONE in the next sentence}

After sampling, the SPAADE-DR dataset is labeled using the most effective prompt and LLMs identified from the evaluation conducted on the Depseverity-Dreaddit dataset. The single-label prompt is applied with LLaMA-3 70b, GPT-4o-mini, and Phi-3.5 MoE to ensure accurate labeling. Since each sample originates from a specific mental disorder subreddit, the original label is retained as the true label for that condition. The remaining five disorders are then annotated using the single-label prompt, ensuring comprehensive multi-label classification. This process allows us to transform the dataset from single-label to multi-label, enabling more comprehensive analysis of co-occurring mental health conditions across the dataset. After labeling, the distribution of each label is calculated to represent the new multi-label characteristics of the dataset, as presented in Table:\ref{tab..exp3}.

\begin{table}[H]
\centering
\caption{The Dataset's Label distribution after conversion to multi-label using Llama-3 70b, GPT-4o mini, and Phi-3.5-MoE}
\label{tab..exp3}
\resizebox{\textwidth}{!}{%
\begin{tabular}{c|ccccccc}
\hline\hline
LLM                          & Label\textbackslash{}Disorder & ADHD & Anxiety & Depression & Eating Disorder & PTSD & Suicide \\ \hline\hline
\multirow{2}{*}{Llama-3 70b} & Positive                      & 1546 & 2547    & 1902       & 564             & 1651 & 971     \\
                             & Negative                      & 1954 & 953     & 1598       & 2936            & 1849 & 2529    \\ \hline
\multirow{2}{*}{Phi-3.5-MoE} & Positive                      & 791 & 3003    & 2190       & 584             & 1496 & 1153    \\
                             & Negative                      & 2709 & 497     & 1310       & 2916            & 2004 & 2347    \\ \hline
\multirow{2}{*}{GPT-4o-mini} & Positive                      & 1783 & 3125    & 2419       & 735             & 1796 & 1477    \\
                             & Negative                      & 1717 & 375     & 1081       & 2765            & 1704 & 2023    \\ \hline\hline
\end{tabular}%
}
\end{table}
% \subsection{}

From Table:\ref{tab..exp3}, it is evident that the label distributions produced by Phi-3.5-MoE are consistently skewed towards either the positive or negative side for all disorders. In contrast, LLaMA-3 70b and GPT-4o-mini exhibit mostly balanced distributions across the disorders.

\section{Additional Analysis}
\label{analysis}
After constructing the multi-label SPAADE-DR dataset, further analyses are conducted to explore associations and comorbidities among various mental disorders, providing deeper insights into their association and comorbidity. Additionally, the performance of multi-label and unrestricted prompts is re-examined to evaluate how they adapt to an increase in the number of labels, from 2 to 6, compared to single-label prompts.

\subsection{Comorbidities between Mental Disorders}
\label{analysis2}
\begin{figure}[htb]
    \centering
    \includegraphics[width=1\linewidth]{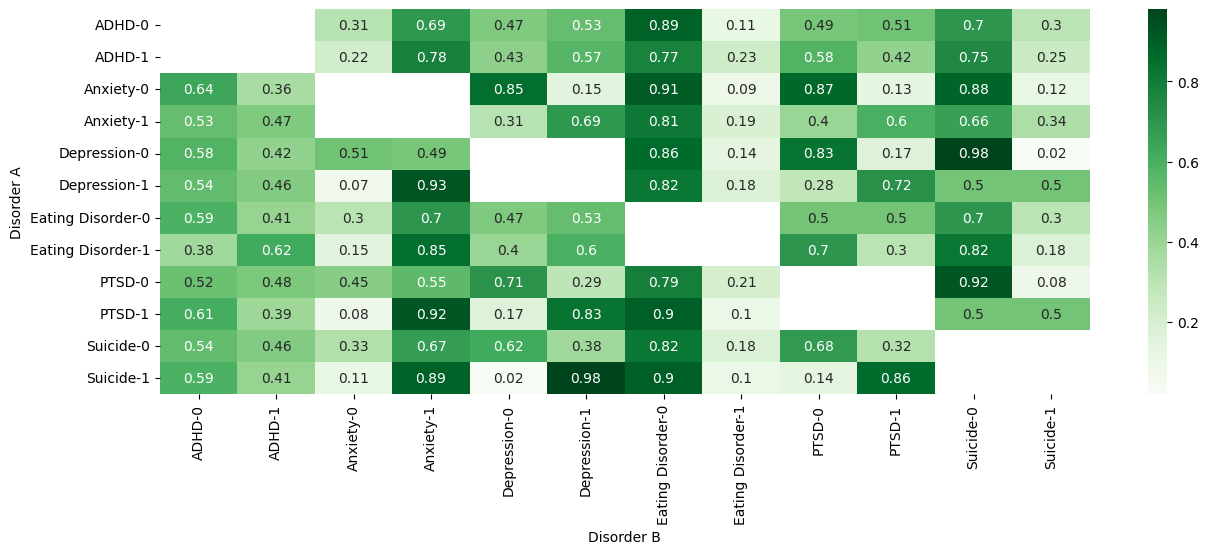}
    \caption{Contingency matrix showing associations between mental disorders (comorbidity)}
    \label{fig:1}
\end{figure}

% To investigate the comorbidities between mental disorders, we use a contingency table (see Fig. \ref{fig:1}) to visualize the associations between primary and secondary disorders. In the table, the y-axis represents the primary mental disorder, and the x-axis represents the secondary disorder. The percentages displayed indicate the proportion of samples with the secondary disorder (positive or negative) among those (positive or negative) for the primary disorder. For instance, the table shows the percentage of samples that are negative for PTSD among those that are positive for suicide.
To explore the comorbidities between mental disorders, a contingency heatmap (as shown in Fig. \ref{fig:1}) is created to visually represent the relationships between 2 disorders, A and B. In this visualization, the y-axis lists the first disorder A, while the x-axis shows the second disorder B. Each cell in the heatmap displays the proportion of samples that exhibit disorder B status (positive or negative) within groups defined by disorder A status (positive or negative). For example, the cell indicating the percentage of samples negative for PTSD among those positive for suicide offers insight into the co-occurrence trends between these two conditions.

This analysis reveals several key insights into the comorbidities between mental disorders:

\begin{itemize}
    \item From figure:\ref{fig:1} it is clear that the likelihood of suicide without depression is low, with only 1.5\% of suicide-positive samples being depression-negative. This suggests a strong association between depression and suicidal tendencies.
    \item Among samples positive for PTSD, only 7.6\% do not also have an anxiety disorder, indicating a high comorbidity rate between PTSD and anxiety.
    \item Depression and anxiety show a strong association, with 93\% of individuals diagnosed with depression also exhibiting symptoms of anxiety. This high comorbidity rate highlights the close relationship between these disorders, which often occur together in clinical settings.
\end{itemize}

\begin{figure}[H]
    \centering
    \includegraphics[width=1\linewidth]{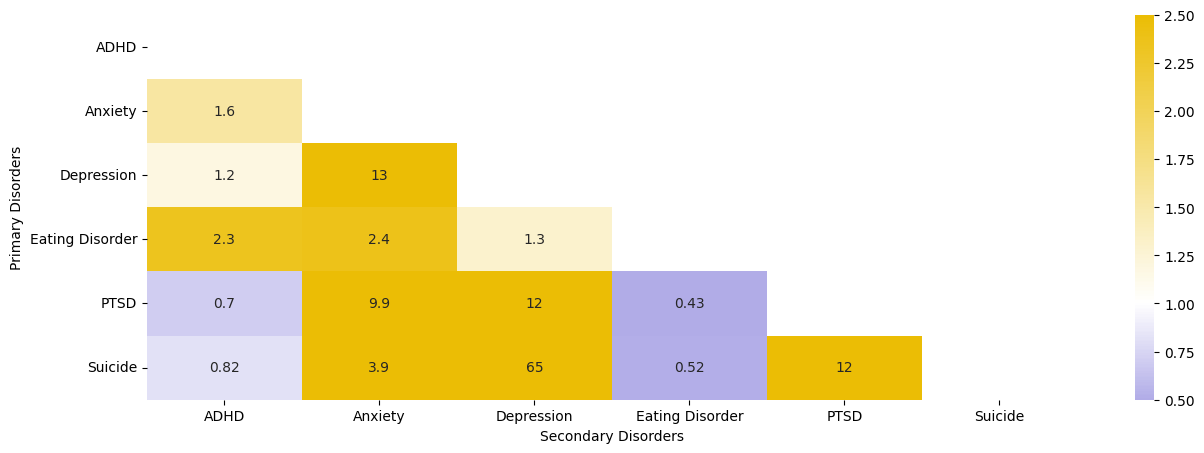}
    \caption{Odds Ratio between mental disorders}
    \label{fig:2}
\end{figure}

The odds ratio (OR) between disorders is then calculated, as depicted in Fig. \ref{fig:2}. The OR serves as a statistical measure of association between two events, commonly used in case-control studies, medical research, and analyses of relationships within datasets. It helps determine how strongly the presence or absence of one event is associated with the presence or absence of another event.

Figure \ref{fig:2} shows that suicide and depression are the two most positively associated disorders, while PTSD and eating disorders are the two most negatively associated disorders. Additionally, the combinations of ADHD \& Suicide and ADHD \& Depression are the least associated disorders.
These comorbidity patterns provide important insights into the co-occurrence of mental disorders and underline the importance of multi-label classification approaches, which can capture these complex relationships within mental health data. This analysis of the association and comorbidity between disorders is aligned with real-world patterns documented in other studies \cite{info:doi/10.2196/jmir.4143, McGrath_Lim_Plana-Ripoll_Holtz_Agerbo_Momen_Mortensen_Pedersen_Abdulmalik_Aguilar-Gaxiola_et_al._2020}.

\subsection{Prompts and LLMs evaluation on 6 disorders}
\label{reeveal}
The multi-label and unrestricted prompts are re-evaluated on the labeled SPAADE-DR dataset to examine how the performance of each model is affected when the number of disorders increases from 2 (as in the Depseverity-Dreaddit dataset) to 6. The evaluation utilizes metrics such as CBA,CF1,OBA,OF1,HL and multi-class BA.

% \usepackage{multirow}
\begin{table}[H]
\centering
\caption{Multi-label and Unrestricted prompts evaluation ana 6 disorders multi-labeled RMHD Dataset}
\label{table_5}
\resizebox{0.999\textwidth}{!}{%
\begin{tabular}{c|c||c|c||c|c||c|c||c|c||c|c||c|c||c|c|c||c}
\hline\hline
 &
  Disorder &
  \multicolumn{2}{c||}{ADHD} &
  \multicolumn{2}{c||}{Anxiety} &
  \multicolumn{2}{c||}{Depression} &
  \multicolumn{2}{c||}{Eating Disorder} &
  \multicolumn{2}{c||}{PTSD} &
  \multicolumn{2}{c||}{Suicide} &
  \multicolumn{3}{c||}{Multi-label} &
  Multi-Class \\ \hline
Prompt                         & LLM\textbackslash{}Metric & BA   & F1   & BA   & F1   & BA   & F1   & BA   & F1   & BA   & F1   & BA   & F1   & GBA  & OF1  & HL   & BA   \\ \hline \hline
\multirow{3}{*}{Multi-Label}   & Llama-3 70b               & 0.51 & 0.06 & 0.63 & 0.63 & 0.67 & 0.63 & 0.51 & 0.04 & 0.61 & 0.45 & 0.71 & 0.58 & 0.64 & 0.50 & 0.33 & 0.09 \\
                               & GPT-4o-mini               & \textbf{0.66} & \textbf{0.48} & \textbf{0.74} & \textbf{0.77 }& \textbf{0.78} & \textbf{0.82} & \textbf{0.96} & \textbf{0.86} & \textbf{0.69} & \textbf{0.57} & \textbf{0.88} & \textbf{0.84} & \textbf{0.78} & \textbf{0.73} & \textbf{0.20} & \textbf{0.16} \\
                               & Phi-3.5-MoE               & 0.51 & 0.07 & 0.62 & 0.68 & 0.59 & 0.65 & 0.89 & 0.85 & 0.55 & 0.45 & 0.69 & 0.55 & 0.65 & 0.56 & 0.33 & 0.08 \\ \hline
\multirow{3}{*}{Unrestricted}  & Llama-3 70b               & 0.50 & 0.02 & 0.68 & 0.81 & 0.69 & 0.71 & 0.47 & 0.08 & 0.57 & 0.43 & 0.74 & 0.61 & 0.66 & 0.58 & 0.32 & 0.06 \\
                               & GPT-4o-mini               & \textbf{0.58} & \textbf{0.28 }& \textbf{0.76} & \textbf{0.88} & \textbf{0.81} & \textbf{0.82} & \textbf{0.80} & \textbf{0.68} & \textbf{0.62} & \textbf{0.40} & \textbf{0.84} & \textbf{0.80} & \textbf{0.76} & \textbf{0.71} & \textbf{0.21} & \textbf{0.13} \\
                               & Phi-3.5-MoE               & 0.50 & 0.02 & 0.64 & 0.77 & 0.61 & 0.57 & 0.64 & 0.38 & 0.53 & 0.35 & 0.68 & 0.54 & 0.62 & 0.52 & 0.35 & 0.08 \\ \hline
\multirow{3}{*}{Single\_Label} & Llama-3 70b               & \textbf{0.97} & \textbf{0.97} & \textbf{1.00} & \textbf{1.00} & \textbf{0.99} & \textbf{0.99} & \textbf{0.98} & \textbf{0.98} & \textbf{0.99} & \textbf{0.99} & \textbf{0.99} & \textbf{0.99} & \textbf{0.99} & \textbf{0.99} & \textbf{0.01} & \textbf{0.90} \\
                               & GPT-4o-mini               & 0.76 & 0.75 & 0.69 & 0.89 & 0.83 & 0.87 & 0.95 & 0.84 & 0.89 & 0.88 & 0.89 & 0.78 & 0.86 & 0.85 & 0.15 & 0.18 \\
                               & Phi-3.5-MoE               & 0.68 & 0.54 & 0.74 & 0.91 & 0.87 & 0.89 & 0.91 & 0.87 & 0.85 & 0.83 & 0.92 & 0.85 & 0.86 & 0.84 & 0.14 & 0.17 \\ \hline \hline
\end{tabular}
}
\end{table}
From Table \ref{table_5}, it is evident that in contrast to the first experiment, where Llama-3 70b emerged as the top-performing model closely followed by GPT-4o-mini and Phi-3.5-MoE, GPT-4o-mini outperforms the other models when the number of disorders in multi-label and unrestricted prompts increases. Meanwhile, the performance of Llama-3 70b and Phi-3.5-MoE significantly declines under these conditions. This indicates that GPT-4o-mini is the most robust and consistent LLM.

Additionally, when using the single-label prompt, Llama-3 70b achieves the highest scores. This is likely due to the alignment of the Llama-3 70b model with the prompt structure used during the data labeling process. Notably, both GPT-4o-mini and Phi-3.5-MoE attain BA and F1 scores above 0.80 across all mental disorders except for ADHD and Anxiety when using the single-label prompt. They both also achieve an overall balanced accuracy (OBA) of 0.86, which is notably high. This suggests that single-label prompts are the most robust, primarily because, despite being computationally more expensive, their performance remains unaffected by the number of labels.

\section{Conclusion \& Future Directions}
\label{conclusion}
This paper explores the potential of large language models (LLMs) to assist in data annotation for mental health research, particularly in the context of social media platforms. Mental health disorders such as depression, anxiety, and PTSD pose significant challenges on a global scale. Social media offer an extensive and valuable source of data that can be harnessed for research into these conditions. Leveraging large language models (LLMs) enables us to improve dataset accuracy, by making them more reflective of real-world data and increasing the efficiency of data annotation processes, thus expanding the volume of high-quality annotated data. This, in turn, can facilitate more effective diagnosis and intervention strategies, ultimately contributing to improved mental health outcomes.

\subsection{Conclusion}
The integration of LLMs into mental health research presents a promising approach for tackling the challenges of large-scale data annotation within this field. The capability of LLMs to process and interpret complex language patterns allows them not only to aid in the identification and diagnosis of mental health conditions through social media analysis but also to enhance the quality and depth of dataset annotations. Notably, LLMs can identify subtle indicators and comorbidities among mental health conditions that traditional methods might miss.

This study leverages LLMs to transition from single-label to multi-label annotation, facilitating more nuanced classification of mental health conditions. Various prompting techniques were tested to identify the most effective strategies, revealing that despite the added complexity of increasing the number of labels or classes, the performance of LLMs remained consistent and robust. These findings highlight the potential of LLMs to revolutionize data annotation processes in mental health research. Additionally, a novel multi-label dataset encompassing six distinct mental disorders was developed, expanding the existing pool of resources and enabling more sophisticated analyses in mental health research.

\subsection{Future Directions}
While LLMs show significant promise, several avenues remain for future exploration:
\begin{itemize}
    \item \textbf{Improving model precision for specific mental health conditions}: LLMs need to be further fine-tuned to detect nuances between different mental health disorders. Customizing models for specific conditions such as depression, anxiety, and severe disorders like psychosis will enhance their diagnostic effectiveness.
    
    \item \textbf{Extending multi-label annotation to other domains}: The multi-label annotation method facilitated by LLMs in this study can be expanded beyond mental health to other domains. Future work should explore how this approach can be applied to various types of data, including text, images, and across different languages.
    
    \item \textbf{Real-time monitoring and intervention}: LLMs could eventually be integrated into real-time monitoring systems, which analyze social media data to detect emerging mental health issues. Such systems could provide timely alerts to health professionals or directly to users when concerning behavioral patterns are identified, potentially preventing crises.
\end{itemize}

In summary, the use of LLMs for mental health data annotation is still in its early stages, but the potential benefits are considerable. Continued refinement of these models, along with addressing current challenges, could position LLMs as a key component in the future of mental health research and interventions.

%Bibliography
% \bibliographystyle{unsrt}  
% \bibliography{references}

\end{document}